\documentclass[a4paper,fleqn]{cas-dc}

\usepackage[numbers]{natbib}

\usepackage{dsfont}
\usepackage{multirow}

\def\tsc#1{\csdef{#1}{\textsc{\lowercase{#1}}\xspace}}
\tsc{WGM}
\tsc{QE}
\tsc{EP}
\tsc{PMS}
\tsc{BEC}
\tsc{DE}

\begin{document}
\let\WriteBookmarks\relax
\def\floatpagepagefraction{1}
\def\textpagefraction{.001}
\shorttitle{Under review as a journal paper at Neurocomputing}
\shortauthors{Song-Lu Chen et~al.}

\title [mode = title]{End-to-end trainable network for degraded license plate detection via vehicle-plate relation mining}                      



\author[1,2]{Song-Lu Chen}[orcid=0000-0002-0780-658X]

\author[1]{Shu Tian}

\author[1,2]{Jia-Wei Ma}

\author[1,2]{Qi Liu}

\author[1,2]{Chun Yang}

\author[2,3]{Feng Chen}

\author[1,2]{Xu-Cheng Yin}
\ead{chenslvs7@gmail.com, xuchengyin@ustb.edu.cn}

\address[1]{School of Computer and Communication Engineering, University of Science and Technology Beijing, Beijing 100083, China}
\address[2]{USTB-EEasyTech Joint Lab of Artificial Intelligence, University of
Science and Technology Beijing, Beijing 100083, China}
\address[3]{EEasy Technology Company Ltd., Zhuhai 519000, China}

\begin{abstract}
License plate detection is the first and essential step of the license plate recognition system and is still challenging in real applications, such as on-road scenarios. In particular, small-sized and oblique license plates, mainly caused by the distant and mobile camera, are difficult to detect. In this work, we propose a novel and applicable method for degraded license plate detection via vehicle-plate relation mining, which localizes the license plate in a coarse-to-fine scheme. First, we propose to estimate the local region around the license plate by using the relationships between the vehicle and the license plate, which can greatly reduce the search area and precisely detect very small-sized license plates. Second, we propose to predict the quadrilateral bounding box in the local region by regressing the four corners of the license plate to robustly detect oblique license plates. Moreover, the whole network can be trained in an end-to-end manner. Extensive experiments verify the effectiveness of our proposed method for small-sized and oblique license plates. Codes are available at: \url{https://github.com/chensonglu/LPD-end-to-end}.
\end{abstract}



\begin{keywords}
license plate detection \sep small-sized license plate \sep oblique license plate \sep end-to-end \sep vehicle-plate relation
\end{keywords}

\maketitle

\section{Introduction}\label{sec1}

License plate detection (LPD) has attracted great interest from academia and industry for many years owing to its importance in many practical applications, such as toll control, parking lot access, and traffic law enforcement. Accurate license plate detection is crucial for subsequent license plate recognition \cite{DBLP:journals/ijon/Wang0QCD18}. However, it remains a challenging task due to illumination variations, background changes, size variations, and viewpoint changes.

Before the deep learning era, most methods \cite{DBLP:journals/ijon/TianSZSZGWZ16,DBLP:journals/tits/AnagnostopoulosAPLK08,DBLP:journals/tcsv/DuISB13,arafat_systematic_2019} need to design handcrafted features for license plate detection. Recently, deep learning methods \cite{chen_simultaneous_2019,DBLP:conf/eccv/SilvaJ18,DBLP:conf/sibgrapi/SilvaJ17} have contributed to improving the license plate detection task. Many methods \citep{chen_simultaneous_2019,DBLP:journals/tip/YuanZZWHK17,DBLP:conf/eccv/XuYMLHYH18,li_towards_2017} propose to localize the license plate directly from the input image, but these methods can not detect small-sized license plates properly, because the license plate is only a small part of the input image. There have been many previous works aiming at small-sized license plate detection by reducing the search area of the license plate using the vehicle proposal \cite{DBLP:conf/eccv/SilvaJ18,DBLP:conf/ijcnn/LarocaSZOGSM18,DBLP:conf/iconip/FuSG17}. However, these methods can not handle large vehicles properly, such as buses and trucks, because their license plates are also a small part of them. Furthermore, \cite{DBLP:conf/sibgrapi/SilvaJ17} presents using the vehicle front region to further reduce the search area of the license plate. The vehicle front region is manually defined as the smallest region comprising the headlights and tires. However, \cite{DBLP:conf/sibgrapi/SilvaJ17} need to manually annotate the location of the vehicle front region, which is ambiguous and a waste of manpower.

Moreover, most previous methods \cite{chen_simultaneous_2019,DBLP:journals/tip/YuanZZWHK17,li_towards_2017,DBLP:conf/eccv/XuYMLHYH18,DBLP:conf/ijcnn/LarocaSZOGSM18,DBLP:conf/sibgrapi/SilvaJ17} simply consider the license plate in horizontal direction, which is only applicable to limited scenes, such as highway bayonet charge and parking lot access. When it comes to more challenging scenes, such as on-road scenarios, they don't work for highly oblique license plates. Although in the literature \cite{DBLP:conf/bmvc/DongHLLZ17,DBLP:conf/eccv/SilvaJ18} there are some methods proposed to detect multi-oriented license plates, they are very complex due to adopting several separate networks.

In this work, we propose an end-to-end trainable network for degraded license plate detection via vehicle-plate relation mining, which can effectively detect the small-sized license plate and accurately localize the quadrilateral bounding box of the oblique license plate in real applications (e.g., on-road scenes). The method can detect the license plate in a coarse-to-fine manner. At the detection stage, we first estimate the location of the license plate based on the offset between the center of the license plate and the vehicle. Considering that the location obtained in this way is not always accurate, we refine the quadrilateral bounding box of the license plate in the local region around the license plate. The local region is simply obtained by expanding the background region around the license plate. In this way, many license plate regions of different vehicles in the input image can be obtained simultaneously, and they have various sizes and aspect ratios. To reduce the running time, all the estimated regions are scaled to the same size and aggregated together into LP patches, so all the license plates can be detected simultaneously in the LP patches. The aforementioned detection stages are combined to build an end-to-end network for license plate detection. Our method can greatly reduce the search area of the license plate, which can minimize false positives and improve the detection performance of small-sized and oblique license plates. Furthermore, estimating the local region can make LPD independent of the size of the vehicles, which is advantageous to large vehicles.

Our main contributions can be summarized as:
\begin{itemize}
\item We propose a novel and applicable method for small-sized and oblique license plate detection by utilizing vehicle-plate relationships, where the license plate is precisely located in a coarse-to-fine scheme. Furthermore, the whole detection network is constructed as an end-to-end trainable network.
\item We propose a novel method to estimate the local region around the license plate via vehicle-plate relation mining, which can greatly reduce the search area of the license plate.
\item We propose a new method to localize the quadrilateral bounding box of the oblique license plate by regressing the four corners of the license plate.
\end{itemize}

The rest of this paper is organized as follows. Related work is described in Section \ref{sec2}. In Section \ref{sec3}, we describe our method in details. Section \ref{sec4} presents comparative experiments and analyses. Final remarks are presented in Section \ref{sec5}.

\section{Related Work}\label{sec2}

\textbf{Direct License Plate Detection}
The following methods propose to localize the license plate directly from the input image. \cite{chen_simultaneous_2019} detects the vehicle and the license plate with two independent branches to remove the effect that the vehicle suppresses the detection of the license plate. \cite{DBLP:journals/tip/YuanZZWHK17} presents a robust and efficient approach for license plate detection, which firstly accelerates the license plate localization using an effective image down-scaling method, then utilizes dense filters to extract candidate regions, and finally identifies the true license plates using a cascaded classifier. \cite{DBLP:conf/eccv/XuYMLHYH18} presents to use multi-scale features to predict and regress the bounding box of the license plate. \cite{li_towards_2017} proposes a method of detecting and recognizing the license plate, where the license plate is localized by Faster R-CNN \cite{DBLP:conf/nips/RenHGS15} directly from the input image. However, these methods can not always detect small-sized license plates properly, because the license plate is only a small part of the input image\footnote{The average size of LPs is 0.26\% of the full input image \cite{DBLP:conf/sibgrapi/SilvaJ17}.}.

\textbf{License Plate Detection with Vehicle Proposal}
The following methods propose to reduce the search area of the license plate by the previous detection of vehicle, vehicle front region, or region around the license plate. In this way, it can improve the detection performance of small-sized license plates and reduce false positives of the license plate. \cite{kim2017deep} utilizes R-CNN\cite{DBLP:conf/cvpr/GirshickDDM14} to generate vehicle proposals and then localizes the license plate in each vehicle region. \cite{DBLP:conf/iconip/FuSG17} applies the Region Proposal Network (RPN) \cite{DBLP:conf/nips/RenHGS15} to generate candidate vehicle proposals and then detects the license plate based on each proposal. \cite{DBLP:conf/eccv/SilvaJ18} introduces a novel CNN framework capable of detecting the license plate in each predicted vehicle region. \cite{DBLP:conf/ijcnn/LarocaSZOGSM18} utilizes YOLOv2 \cite{DBLP:conf/cvpr/RedmonF17} to detect all the vehicles, then localizes the license plates in the vehicle patches simultaneously. However, these methods are still not favorable to large vehicles, such as trucks and buses, because their license plates are still only a small part of them. \cite{DBLP:conf/sibgrapi/SilvaJ17} proposes to detect the vehicle first, then detects the vehicle front region, and finally localizes the license plate in each vehicle front region. However, the vehicle front region needs to be annotated manually, which is a waster of manpower. \cite{DBLP:journals/tits/XieAJLZ18} employs an attention-like method to estimate the local region around the license plate, then detects the license plate in the local region. However, in the literature \cite{DBLP:journals/tits/XieAJLZ18}, it utilizes low-resolution feature map and ROI pooling \citep{DBLP:conf/iccv/Girshick15} for LPD, which causes loss of the spatial and semantic information, so the end-to-end model in \citep{DBLP:conf/iccv/Girshick15} suffers significant performance degradation, especially for the large IOU threshold. Our method can maintain the semantic information by adopting the high-resolution feature map, and retain the spatial information by using space-invariant ROI warping \cite{DBLP:conf/cvpr/DaiHS16}.

\textbf{Multi-Oriented License Plate Detection}
\cite{DBLP:journals/tits/XieAJLZ18} proposes a CNN-based MD-YOLO framework for multi-directional license plate detection via rotation angle prediction. \cite{DBLP:journals/sensors/HanYZTL19} proposes to detect the license plate with a tightly bounding parallelogram via predicting the top-left, top-right and bottom-right corners of the license plate. \cite{DBLP:journals/jei/TianWL19} utilizes semantic segmentation to get the rotation angle of the license plate. \cite{DBLP:journals/tits/XieAJLZ18,DBLP:journals/sensors/HanYZTL19,DBLP:journals/jei/TianWL19} regard the oblique license plate as a parallelogram. However, it is not always accurate due to the perspective transformation of the license plate, because a highly oblique license plate is more like a arbitrary quadrilateral. \cite{DBLP:conf/bmvc/DongHLLZ17} proposes to generate license plate candidates with RPN \cite{DBLP:conf/nips/RenHGS15}, then uses R-CNN \cite{DBLP:conf/cvpr/GirshickDDM14} to regress the four corners of the license plate. \cite{DBLP:conf/eccv/SilvaJ18} proposes to obtain the affine transformation parameters explicitly based on Spatial Transformer Networks (STN) \cite{DBLP:conf/nips/JaderbergSZK15}, which can transform the tilted license plate into a horizontal direction. However, \cite{DBLP:conf/bmvc/DongHLLZ17,DBLP:conf/eccv/SilvaJ18} are complicated due to adopting several separate networks. Our method can localize the four corners of the license plate in an end-to-end manner.

\textbf{Vehicle Detection}
Before the deep learning era, most methods \cite{DBLP:journals/pami/SunBM06} usually utilized information about symmetry, color, shadow, geometrical features (e.g., corners, horizontal/vertical edges), texture features and vehicle lights for vehicle detection. Recently, deep learning methods \cite{DBLP:journals/ijon/CaoJCW16,DBLP:journals/ijon/MoHZXQ19} have contributed to improving vehicle detection. \cite{DBLP:journals/sensors/WangLD19,DBLP:journals/sensors/SangWGHXZC18,DBLP:journals/sensors/WuSZXCX19} design better anchor priors for vehicle detection, which can facilitate the matching between the anchor box and the ground truth box. \cite{DBLP:journals/tits/HuXXCHQH19,DBLP:conf/avss/KimKCC18} utilize multi-scale features to be robust to scale change of the vehicles by adopting YOLOv3 \cite{DBLP:journals/corr/abs-1804-02767}. \cite{DBLP:conf/icmcs/LiuLHLZ18,DBLP:journals/ijon/LiuLH19} introduce a backward feature enhancement network to generate high-recall proposals, then adopt a spatial layout preserving network to enhance tiny vehicle detection. In this work, we simply adopt vanilla SSD \cite{DBLP:conf/eccv/LiuAESRFB16} for vehicle detection, which can detect various-sized vehicles by utilizing multi-scale features. We will employ more powerful vehicle detectors in future work.

\begin{figure*}
    \centering
        \includegraphics[width=1.0\linewidth]{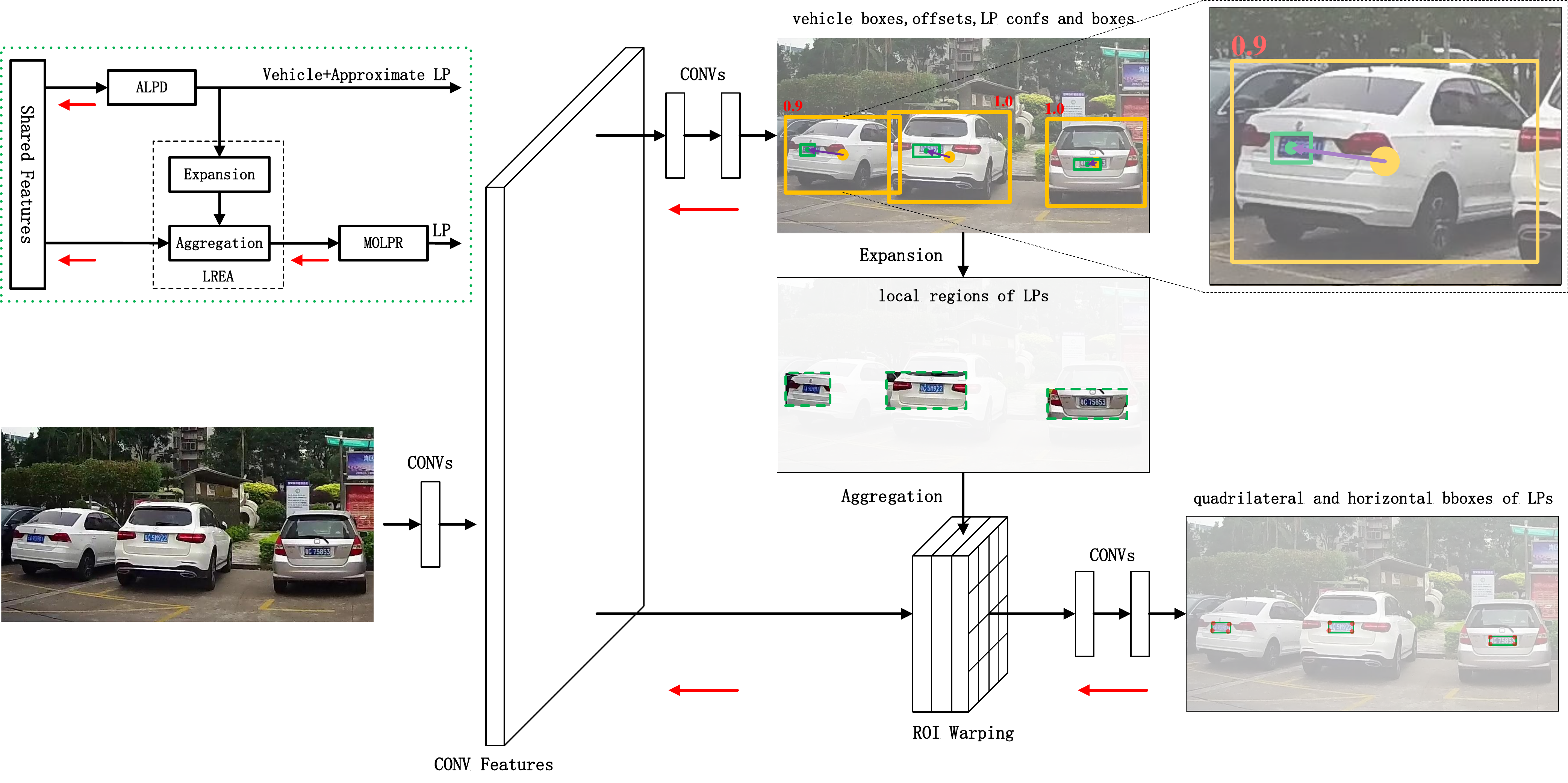}
    \caption{A thumbnail of the overall architecture is shown in the top-left corner (ALPD: approximate license plate detection; LREA: local region estimation and aggregation; MOLPR: multi-oriented license plate refinement). At the \textbf{ALPD} stage, first, the vehicle (orange rectangle) is detected, so the center of the vehicle (orange circle) is determined; second, the center of the license plate (green circle) is obtained based on the offset (purple arrow) between the center of the license plate and the vehicle; third, the size of the license plate is directly predicted from the input image. According to the center and size of the license plate, we can approximately estimate the license plate (green rectangle). Moreover, the probability of the vehicle containing a license plate (red number) is predicted simultaneously. An enlarged example is shown in the top-right corner. At the \textbf{LREA} stage, the local region of LP is obtained by expanding the background region around the license plate with a preset ratio, then all the expanded LP regions (green dashed rectangle) are aggregated into feature patches via ROI warping \cite{DBLP:conf/cvpr/DaiHS16}, in which all the LP regions have the same size and aspect ratio. Therefore, all the license plates can be detected simultaneously in the feature patches. At the \textbf{MOLPR} stage, the quadrilateral (red circle) and horizontal (green rectangle) bounding boxes of the license plate are detected simultaneously in the local region of LP. The network can be trained in an end-to-end manner, where the red arrows denote the backpropagation gradients.}
    \label{fig:architecture}
\end{figure*}

\section{Methodology}\label{sec3}
We propose an end-to-end trainable network for degraded license plate detection via vehicle-plate relation mining, which detects the license plate in a coarse-to-fine manner. The overall architecture is illustrated in Figure \ref{fig:architecture}, where it firstly predicts the approximate location of the license plate utilizing spatial relationships between the license plate and the vehicle (Section \ref{sec3.1}), then estimates the local region by expanding the background region around the license plate followed by an aggregation operation (Section \ref{sec3.2}), and finally refines the quadrilateral bounding box of the license plate (Section \ref{sec3.3}).

\subsection{Approximate License Plate Detection}\label{sec3.1}

At this stage, the vehicle is firstly detected, so the center of the vehicle is determined. After that, the network predicts the approximate location and size of the license plate, where the location is obtained based on the offset between the center of the license plate and the vehicle, and the size is directly predicted from the input image. Besides, the probability of the vehicle containing a license plate is predicted simultaneously. As shown in Figure \ref{fig:architecture}, the location, and the size of the license plate are not accurate in general cases, because they are directly predicted from the large input image, of which the license plate is only a small portion.

\begin{table}[width=.9\linewidth,cols=4,pos=h]
\caption{Backbone network of the ALPD network.}\label{tab:backbone-1}
\begin{tabular*}{\tblwidth}{@{} LLLL@{} }
\toprule
Type                & Filters & Parameters  & Output  \\
\midrule
Convolution$\Delta$   & 64      & k:3,s:1     & 512 $\times$ 512 \\
Convolution   & 64      & k:3,s:1     & 512 $\times$ 512 \\
Maxpool             & -       & k:2,s:2     & 256 $\times$ 256 \\
Convolution   & 128     & k:3,s:1     & 256 $\times$ 256 \\
Convolution   & 128     & k:3,s:1     & 256 $\times$ 256 \\
Maxpool             & -       & k:2,s:2     & 128 $\times$ 128 \\
Convolution   & 256     & k:3,s:1     & 128 $\times$ 128 \\
Convolution   & 256     & k:3,s:1     & 128 $\times$ 128 \\
Convolution   & 256     & k:3,s:1     & 128 $\times$ 128 \\
Maxpool             & -       & k:2,s:2     & 64 $\times$ 64   \\
Convolution   & 512     & k:3,s:1     & 64 $\times$ 64   \\
Convolution   & 512     & k:3,s:1     & 64 $\times$ 64   \\
Convolution*  & 512     & k:3,s:1     & 64 $\times$ 64   \\
Maxpool             & -       & k:2,s:2     & 32 $\times$ 32   \\
L2Norm              & -       & -           & 32 $\times$ 32   \\
Convolution   & 512     & k:3,s:1     & 32 $\times$ 32   \\
Convolution   & 512     & k:3,s:1     & 32 $\times$ 32   \\
Convolution   & 512     & k:3,s:1     & 32 $\times$ 32   \\
Maxpool             & -       & k:3,s:1     & 32 $\times$ 32   \\
Convolution    & 1024    & k:3,s:1,d:6 & 32 $\times$ 32   \\
Convolution*   & 1024    & k:1,s:1     & 32 $\times$ 32   \\
Convolution   & 256     & k:1,s:1     & 32 $\times$ 32   \\
Convolution*  & 512     & k:3,s:2     & 16 $\times$ 16   \\
Convolution   & 128     & k:1,s:1     & 16 $\times$ 16   \\
Convolution*  & 256     & k:3,s:2     & 8 $\times$ 8     \\
Convolution   & 128     & k:1,s:1     & 8 $\times$ 8     \\
Convolution*  & 256     & k:3,s:2     & 4 $\times$ 4     \\
Convolution   & 128     & k:1,s:1     & 4 $\times$ 4     \\
Convolution*  & 256     & k:3,s:2     & 2 $\times$ 2     \\
Convolution  & 128     & k:1,s:1     & 2 $\times$ 2     \\
Convolution* & 256     & k:4,s:1     & 1 $\times$ 1     \\
\bottomrule
\end{tabular*}
\end{table}

The ALPD network is based on SSD \cite{DBLP:conf/eccv/LiuAESRFB16} for multi-task learning. The backbone network is shown in Table \ref{tab:backbone-1} without showing the ReLU activation function, and it is transformed from VGG-16 \cite{simonyan2015very} followed by several extra layers. As for parameters, "k, s, d" mean kernel size, stride size, and dilation parameters \cite{DBLP:journals/corr/YuK15} respectively. Moreover, we apply L2Norm \cite{DBLP:journals/corr/LiuRB15} before combining the shallow features and the deep features, which avoids that the larger parameters "dominate" the smaller ones. As mentioned in Section \ref{sec3.3}, the first convolutional layer marked with "$\Delta$" is the input layer of the MOLPR stage. Besides, layers marked with "*" are candidates for multi-scale detection.

The training objective of the ALPD network is defined as (\ref{eq:coarse-loss}), which consists of the classification loss of the vehicle $ L_{conf}(c)$, the regression loss of the vehicle $L_{loc}(l,g)$, the loss of whether the vehicle contains a license plate $L_{has\_lp}(\upsilon,lpc)$, the loss of the offset between the center of the licence plate and the vehicle $L_{off}(l,g,\upsilon)$, and the size loss of the license plate $L_{lp_{wh}}(l,g,\upsilon)$:

\begin{equation}
\begin{split}
\label{eq:coarse-loss}
    L_1(c,l,g,\upsilon,lpc)=\frac1{N}\lbrack
    L_{conf}(c)+L_{loc}(l,g)\\
    +L_{has\_lp}(\upsilon,lpc)+L_{off}(l,g,\upsilon)+L_{lp_{wh}}(l,g,\upsilon)\rbrack
\end{split}
\end{equation}
where $N$ is the number of the matched default boxes with the ground truth boxes of the vehicle, $c$ is the confidence of the vehicle, $l$ is the predicted box of the vehicle, $g$ is the ground truth box of the vehicle, $\upsilon$ is the ground truth of whether the vehicle contains a license plate, and $lpc$ is the predicted probability of the vehicle containing a license plate.

The learning target of vehicle detection is completely consistent with SSD \cite{DBLP:conf/eccv/LiuAESRFB16}, which predicts the vehicle presence confidence with cross-entropy loss (\ref{eq:conf-loss}) and regresses the bounding box of the vehicle with smooth L1 loss (\ref{eq:loc-loss}):

\begin{equation}
\label{eq:conf-loss}
    L_{conf}(c)=-\sum_{i=1}^N\sum_p\log(c_i^p)\;\;\;\;\;\;c_i^p=\frac{exp(c_i^p)}{\displaystyle{\textstyle\sum_{}}exp(c_i^p)}
\end{equation}

\begin{equation}
\label{eq:loc-loss}
\resizebox{.91\linewidth}{!}{$
    \displaystyle
    L_{loc}(l,g)=\sum_{i=1}^{N}\sum_{m\in\left\{cx,cy,w,h\right\}}\mathds{1}_{ij}^{p^+}smooth_{L1}\left(l_i^m-\overline g_j^m\right)
$}
\end{equation}

\begin{equation}
\label{eq:smoothl1}
smooth_{L1}=\left\{\begin{array}{l}0.5x^2\;\;\;\;\;\;\;\;\;\left|x\right|<1\\\left|x\right|-0.5\;\;\;\;otherwise\end{array}\right.
\end{equation}
where the category $p$ is $\{vehicle, background\}$, the positive category $p^+$ is $vehicle$, and $\mathds{1}_{ij}^{p^+}\in\left\{0,1\right\}$ is the indicator of whether the $i$-th default box matches the $j$-th ground truth box. The smooth L1 loss \cite{DBLP:conf/iccv/Girshick15} is defined as (\ref{eq:smoothl1}). Similar to SSD \cite{DBLP:conf/eccv/LiuAESRFB16}, the vehicle is regressed based on the center $(cx, cy)$ of the matched default box $(d)$ and its width $(w)$ and height $(h)$, as shown in (\ref{eq:loc-xywh-loss}).

\begin{equation}
\begin{split}
\label{eq:loc-xywh-loss}
    \overline g_j^{cx}={\left(g_j^{cx}-d_i^{cx}\right)/{d_i^w}}\;\;\;\;\;\;\;\;\overline g_j^{cy}={\left(g_j^{cy}-d_i^{cy}\right)/{d_i^h}}\\
    \overline g_j^w=\log\left(g_j^w/{d_i^w}\right)\;\;\;\;\;\;\;\;\overline g_j^h=\log\left(g_j^h/{d_i^h}\right)
\end{split}
\end{equation}

The probability of whether the vehicle contains a license plate can reduce false positives of the license plate. Very small-sized vehicles and vehicles with invisible license plate (occlusion, large vehicle pose, etc.) are recognized as without license plate; otherwise, the vehicles are labeled as containing a license plate. The probability is optimized by binary cross-entropy loss (\ref{eq:has-lp-loss}), where $\sigma$ is a sigmoid function to limit the confidence to $\left[0,1\right]$.

\begin{equation}
\begin{split}
\label{eq:has-lp-loss}
    L_{has\_lp}(\upsilon,lpc)=-\sum_{i=1}^{N}\lbrack
    \upsilon_i\cdot\log(\sigma\left(lpc_i\right))\\
    +\left(1-\upsilon_i\right)\cdot\log\left(1-\sigma\left(lpc_i\right)\right)\rbrack
\end{split}
\end{equation}

The offset between the center of the license plate and the vehicle as well as the size of the license plate are estimated with smooth L1 loss \cite{DBLP:conf/iccv/Girshick15}, as shown in (\ref{eq:offset-size-loss}). The vehicle should contain a license plate; otherwise, the losses $L_{off}(l,g,\upsilon)$ and $L_{lp_{wh}}(l,g,\upsilon)$ are all set to 0 by setting $\upsilon_i=0$:

\begin{equation}
\label{eq:offset-size-loss}
\resizebox{.91\linewidth}{!}{$
    \displaystyle
    \begin{split}
    L_{off}(l,g,\upsilon)=\sum_{i=1}^{N}\sum_{m\in\left\{off_{x,y}\right\}}\mathds{1}_{ij}^{p^+}\upsilon_ismooth_{L1}\left(l_i^m-\overline g_j^m\right)\\
    L_{lp_{wh}}(l,g,\upsilon)=\sum_{i=1}^{N}\sum_{m\in\left\{lp_{w,h}\right\}}\mathds{1}_{ij}^{p^+}\upsilon_i smooth_{L1}\left(l_i^m-\overline g_j^m\right)
    \end{split}
$}
\end{equation}
where $off_{x,y}$ is the offset between the center of the license plate and the vehicle in both x-direction and y-direction, and $lp_{w,h}$ is the width and height of the license plate. Based on the matched default box, it directly regresses the offset and limits the LP size to $\left(0,+\infty\right)$ by a logarithmic operation to prevent negative numbers, as shown in (\ref{eq:offxylpwd-loss}).

\begin{equation}
\begin{split}
\label{eq:offxylpwd-loss}
    \overline g_j^{off_x}=g_j^{off_x}/{d_i^w}\;\;\;\;\;\;\;\;\overline g_j^{off_y}=g_j^{off_y}/{d_i^h}\\
    \overline g_j^{lp_w}=\log\left(g_j^{lp_w}/{d_i^w}\right)\;\;\;\;\;\;\;\;\overline g_j^{lp_h}=\log\left(g_j^{lp_h}/{d_i^h}\right)
\end{split}
\end{equation}

\subsection{Local Region Estimation and Aggregation}\label{sec3.2}

Compared with detecting the license plate directly in the large input image, it is better to localize the license plate in a small local region. Based on the predicted center and size of the license plate obtained at the ALPD stage, the local region can be obtained by simply expanding the background region around the license plate with a preset ratio, which can not exceed the boundary of the corresponding vehicle to reduce redundant background noises.

After that, many license plate regions of different vehicles can be obtained simultaneously. However, these LP regions have different sizes and aspect ratios. To reduce the running time, all the estimated regions are aggregated via ROI warping \cite{DBLP:conf/cvpr/DaiHS16} as feature patches, in which all components are scaled to the same size and aspect ratio. Therefore, all the license plates can be detected simultaneously in the feature patches. All the LP region features are extracted from the first convolutional layer of the ALPD network, as seen in Table \ref{tab:backbone-1}, because the first convolutional layer has the same size as the input image, which preserves the spatial information and is favorable to the detection of small-sized license plates.

\subsection{Multi-Oriented License Plate Refinement}\label{sec3.3}

At this stage, the quadrilateral and horizontal bounding boxes of the license plate are detected simultaneously in the local region around the license plate. The detection results are more accurate than those obtained at the ALPD stage, as illustrated in Figure \ref{fig:architecture}.

\begin{table}[width=.9\linewidth,cols=4,pos=h]
\caption{Backbone network of the MOLPR network.}\label{tab:backbone-2}
\begin{tabular*}{\tblwidth}{@{} LLLL@{} }
\toprule
Type         & Filters & Parameters & Output \\
\midrule
Convolution  & 512     & k:3,s:1    & 56 $\times$ 56  \\
Convolution* & 512     & k:3,s:1    & 56 $\times$ 56  \\
Maxpool      & -       & k:2,s:2    & 28 $\times$ 28  \\
Convolution  & 512     & k:3,s:1    & 28 $\times$ 28  \\
Convolution* & 512     & k:3,s:1    & 28 $\times$ 28  \\
Maxpool      & -       & k:2,s:2    & 14 $\times$ 14  \\
Convolution  & 512     & k:3,s:1    & 14 $\times$ 14  \\
Convolution* & 512     & k:3,s:1    & 14 $\times$ 14  \\
\bottomrule
\end{tabular*}
\end{table}

The backbone network of this stage is shown in Table \ref{tab:backbone-2}, where ``k'' means kernel size and ``s'' means stride size. Like SSD \cite{DBLP:conf/eccv/LiuAESRFB16}, layers marked by ``*'' are candidates for multi-scale detection.

The training objective of the MOLPR network is defined as (\ref{eq:refine-loss}), which consists of the classification loss of the horizontal license plate $L_{conf}(c')$, the regression loss of the horizontal license plate $L_{loc}(l',g')$, and the corner loss of the quadrilateral license plate $L_{corner}(l',g')$:

\begin{equation}
\begin{split}
\label{eq:refine-loss}
    L_2(c',l',g')=\frac1{N'}\lbrack L_{conf}(c')+L_{loc}(l',g')\\
    +L_{corner}(l',g')\rbrack
\end{split}
\end{equation}
where $N'$ is the number of the matched default boxes with the ground truth boxes of the license plate, $c'$ is the confidence of the license plate, $l'$ is the predicted box of the license plate, and $g'$ is the ground truth box of the license plate. The losses of the horizontal license plate $L_{conf}(c')$ and $L_{loc}(l',g')$ are similar to vehicle detection, as shown in (\ref{eq:conf-loss}) and (\ref{eq:loc-loss}).

\begin{figure}
    \centering
        \includegraphics[width=.9\linewidth]{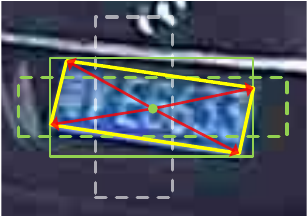}
    \caption{Four corners regression (red arrow) of the license plate. The matched default box (green dashed) is responsible for regressing the quadrilateral bounding box (yellow solid), where the default box is evaluated by IOU with the horizontal ground truth box (green solid). The four corners of the license plate are obtained based on the center of the matched default box. The irrelevant default box (grey dashed) is ignored because of low IOU.}
    \label{fig:four-corners}
\end{figure}

The quadrilateral bounding box of the license plate is obtained by regressing the four corners of the license plate, as illustrated in Figure \ref{fig:four-corners}. The corner loss of the license plate is optimized with smooth L1 loss \cite{DBLP:conf/iccv/Girshick15}, as shown in (\ref{eq:lp-corners-loss}):

\begin{equation}
\label{eq:lp-corners-loss}
\resizebox{.9\linewidth}{!}{$
    \displaystyle
    L_{corner}(l',g')=\sum_{i=1}^{N'}\sum_{m\in\left\{tl,tr,br,bl\right\}}\mathds{1}_{ij}^{{p'}^+} smooth_{L1}\left({l'}_i^m-\overline {g'}_j^m\right)
$}
\end{equation}
where the positive category ${p'}^+$ is $license\;plate$ and $m\in\left\{tl,tr,br,bl\right\}$ are the four corners (top-left, top-right, bottom-right, bottom-left) of the license plate. The regression target is shown in (\ref{eq:lp-corners-xy-loss}), where the four corners of the license plate are directly regressed based on the center ($cx$, $cy$) of the matched default box ($d'$) and its width ($w$) and height ($h$).

\begin{equation}
\begin{split}
\label{eq:lp-corners-xy-loss}
    \overline {g'}_j^X=\left({g'}_j^X-{d'}_i^{cx}\right)/{{d'}_i^w}\;\;\;\;X\in\left\{tlx,trx,brx,blx\right\}\\
    \overline {g'}_j^Y=\left({g'}_j^Y-{d'}_i^{cy}\right)/{{d'}_i^h}\;\;\;\;Y\in\left\{tly,try,bry,bly\right\}
\end{split}
\end{equation}

\subsection{End-to-End Trainable Detection Network}\label{sec3.4}

By integrating the aforementioned detection stages, we develop an end-to-end trainable network for degraded license plate detection. Combining (\ref{eq:coarse-loss}) and (\ref{eq:refine-loss}) together, the loss of the whole network is shown in (\ref{eq:total-loss}), where $\alpha$ is simply set to 1 to balance these loss terms.

\begin{equation}
\label{eq:total-loss}
    L=L_{1}(c,l,g,\upsilon,lpc)+\alpha L_{2}(c',l',g')
\end{equation}

\begin{figure}
    \centering
        \includegraphics[width=.9\linewidth]{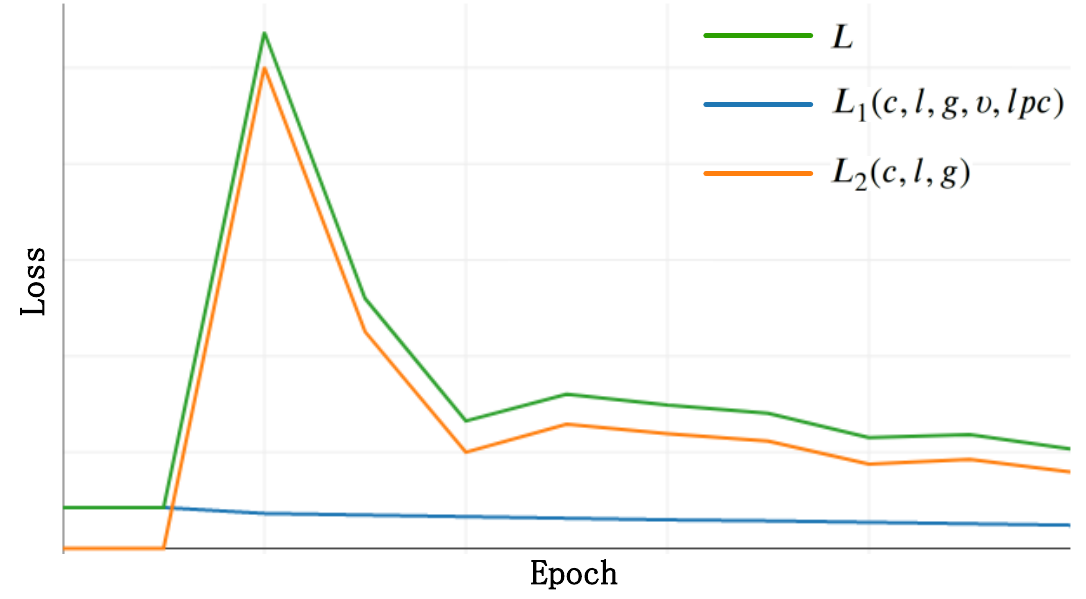}
    \caption{Training loss. $L_{1}(c,l,g,\upsilon,lpc)$ is the loss of the ALPD network, $L_{2}(c',l',g')$ is the loss of the MOLPR network, and $L$ is the total loss of the end-to-end network.}
    \label{fig:total-loss}
\end{figure}

During end-to-end training, the ALPD network can be firstly optimized to estimate the local region of the license plate, then the entire network will be optimized simultaneously. Concretely, during the first few epochs, $L_{1}$ goes down and $L_{2}$ remains zero because the untrained ALPD network can not estimate the location of the license plate; then $L_{2}$ goes up dramatically because the ALPD network can approximately estimate the license plate after training for some epochs, and the MOLPR network starts learning to regress the four corners of the license plate; finally the total loss $L$ goes down steadily because the ALPD and MOLPR network are optimized simultaneously.

\begin{table*}[width=2.0\linewidth,cols=8,pos=h]
\caption{Ablation study (AP) of different  datasets and IOU thresholds.}\label{tab:ablation}
\begin{tabular*}{\tblwidth}{@{} LCCCCCCC@{} }
\toprule
\multirow{2}{*}{Method} & \multirow{2}{*}{LREA \& MOLPR} & \multirow{2}{*}{has\_lp confidence} & \multirow{2}{*}{vehicle boundary} & \multicolumn{2}{c}{IOU=0.5}                                                                                      & \multicolumn{2}{c}{IOU=0.75}                                                                                     \\ \cline{5-8} 
                        &                               &                                     &                                   & \begin{tabular}[c]{@{}c@{}}TILT720\\ Test\end{tabular} & \begin{tabular}[c]{@{}c@{}}TILT1080\\ Test\end{tabular} & \begin{tabular}[c]{@{}c@{}}TILT720\\ Test\end{tabular} & \begin{tabular}[c]{@{}c@{}}TILT1080\\ Test\end{tabular} \\
\midrule
Ours (ALPD)           &                               &                                     &                                   & 76.71\%                                                & 77.71\%                                                 & 26.27\%                                                & 35.27\%                                                 \\
\midrule
\multirow{3}{*}{Ours (E2E)}   & $\surd$                             &                                     &                                   & 88.11\%                                                & 86.20\%                                                 & 53.73\%                                                & 53.37\%                                                 \\
                        & $\surd$                             & $\surd$                                   &                                   & 88.95\%                                                & 87.29\%                                                 & 54.46\%                                                & 55.94\%                                                 \\
                        & $\surd$                             & $\surd$                                   & $\surd$                                 & \textit{\textbf{89.19\%}}                              & \textit{\textbf{87.67\%}}                               & \textit{\textbf{54.51\%}}                              & \textit{\textbf{56.92\%}}                               \\ \bottomrule
\end{tabular*}
\end{table*}

\section{Experiments}\label{sec4}
We mainly follow SSD \cite{DBLP:conf/eccv/LiuAESRFB16}, including the data augmentation strategies, such as random crop and color distortion, etc. We adopt SSD512 as the baseline network of the ALPD stage, which is initialized with the ILSVRC CLS-LOC dataset \cite{DBLP:journals/ijcv/RussakovskyDSKS15}. The backbone network of the MOLPR stage is trained from scratch, where the input size is $56 \times 56 \times 64$ (height $\times$ width $\times$ channel). Our model is trained for 60K iterations using the Adam optimizer \cite{DBLP:journals/corr/KingmaB14}. The momentum parameters are set to $\beta_1 = 0.9$ and $\beta_2 = 0.999$. Batch size and weight decay are set to $32$ and $5 \times 10^{-4}$ respectively. The learning rate is first initialized to $10^{-4}$ and then decreased 10 times at 20K and 40K iterations. All the experiments are carried on a PC with 4 NVIDIA Titan Xp GPUs.

\subsection{Datasets}\label{sec4.1}

\textbf{TILT720.}
We employ an automobile data recorder to collect on-road videos with a size of $720 \times 1280$. After keyframe extraction and careful annotation, a total of 1033 images are obtained. All visible vehicles are labeled with a horizontal bounding box, and their corresponding license plates are annotated with a quadrilateral bounding box. The horizontal bounding box of the license plate is regarded as the tightest boundary of the quadrilateral bounding box. For simplicity, we name the dataset TILT720 (mulTi-oriented lIcense pLate deTection dataset 720p). All images are randomly divided into the training-validation set and test set by 9:1.

\textbf{TILT1080.}
Similar to the TILT720 dataset, we obtain the TILT1080 dataset with another automobile data recorder. The TILT1080 dataset contains 4112 images, and all images have a size of $1080 \times 1920$. All images are randomly divided into the training-validation set and test set by 9:1.

\subsection{Evaluation Protocols}\label{sec4.2}
\textbf{Horizontal Bounding Box.}
We adopt the general AP (Average Precision) to evaluate the horizontal bounding box of the license plate. To be specific, we follow the 11-points computation of the VOC2007 \cite{DBLP:journals/ijcv/EveringhamGWWZ10} with different IOU thresholds (0.5 and 0.75). If it is not specified, the IOU threshold is set to 0.5.

Moreover, we hope that the expanded region at the LREA stage could completely contain the license plate for the next MOLPR stage. To evaluate it, we define a new evaluation criterion $C_{recall}$, as shown in (\ref{eq:c-recall}):

\begin{equation}
\label{eq:c-recall}
    C_{recall}=\frac{\displaystyle1}M\sum_{i=1}^MER_i\cap LP_i=LP_i
\end{equation}
where $ER_i$ means the $i$-th expanded region, $LP_i$ denotes the $i$-th license plate, and $M$ is the number of LP ground truths.

\textbf{Quadrilateral Bounding Box.}
We adopt the classical precision (P), recall (R) and $F_1$-score (F) to evaluate the quadrilateral bounding box of the license plate:

\begin{equation}
\label{eq:PRF}
    P=\frac{TP}{TP+FP}\;\;\;\;R=\frac{TP}{TP+FN}\;\;\;\;F=\frac{2PR}{P+R}
\end{equation}
where TP means true positive, FP means false positive, and FN means false negative. A quadrilateral bounding box is considered as correct when the IOU with a quadrilateral ground truth box is greater than the threshold (0.5 or 0.75) under the confidence threshold 0.5.

\subsection{Ablation Study}\label{sec4.3}

As demonstrated in Table \ref{tab:ablation}, we adopt the ALPD network as the baseline model. The network only achieves 26.27\% and 35.27\% on the test set of TILT720 and TILT1080 with IOU threshold 0.75, because it fails to accurately localize the license plate from the large input image.

\textbf{LREA \& MOLPR.}
By adding the LREA stage and MOLPR stage, we obtain the end-to-end detection network, as illustrated in Figure \ref{fig:architecture}. With a large IOU threshold, it improves almost 20\% on the test set because of localizing the license plate in the local region, which proves the effectiveness of our method. With a small IOU threshold, it can also improve about 10\% on the test set.

\textbf{has\_lp confidence.}
The ALPD network will inevitably estimate the approximate location and size of the license plate, no matter there is a visible license plate or not. However, the license plate is not always visible, especially for the very small-sized vehicle, occluded vehicle, and large-posed vehicle. The confidence of whether the vehicle contains a license plate can reduce false positives of the license plate, and it is fixed to 0.5. In this way, the invisible license plate can be filtered out and the precision is improved.

\textbf{vehicle boundary.}
The expanded region at the LREA stage can be limited by the predicted vehicle boundary to avoid redundant background noises. It can further improve the performance on both test sets with different IOU thresholds.

\begin{table}[width=.8\linewidth,cols=4,pos=h]
\caption{The influence to vehicle detection (AP) with IOU threshold 0.5. "E2E" means end-to-end.}\label{tab:ablation-vehicle}
\begin{tabular*}{\tblwidth}{@{} LLL@{} }
\toprule
Method         & \begin{tabular}[c]{@{}c@{}}TILT720\\ Test\end{tabular} & \begin{tabular}[c]{@{}c@{}}TILT1080\\ Test\end{tabular} \\
\midrule
SSD            & 87.82\%                                                & 87.51\%                                                 \\
Ours (ALPD) & 87.85\%                                                & 87.50\%                                                 \\
Ours (E2E)     & 87.83\%                                                & 87.52\%                                                 \\
\bottomrule
\end{tabular*}
\end{table}

Moreover, as demonstrated in Table \ref{tab:ablation-vehicle}, either the ALPD network or the end-to-end network (E2E), our methods have no influence on vehicle detection compared with vanilla SSD \cite{DBLP:conf/eccv/LiuAESRFB16}. As can be seen, there is still a lot of room to improve the performance of vehicle detection, and we will employ more powerful vehicle detectors in future work.

\subsection{Experiments with Expansion Ratio}\label{sec4.4}

We hope the expanded region, obtained at the LREA stage, can fully contain the license plate for the next MOLPR stage. The expanded region is obtained based on the center and size of the license plate predicted at the ALPD stage, which takes the center of the license plate as the center and expands the width and height with the same ratio. To verify the effect of different expansion ratios at the LREA stage, we conduct comparative experiments on the trainval set and test set of TILT720. Apart from the expansion ratio at the LREA stage, all other settings of our end-to-end network are the same.

\begin{table}[width=.9\linewidth,cols=5,pos=h]
\caption{The effect of different expansion ratios at the LREA stage.}\label{tab:expand-num}
\begin{tabular*}{\tblwidth}{@{} LCCCC@{} }
\toprule
\multirow{2}{*}{\begin{tabular}[c]{@{}l@{}}Expansion\\ Ratio\end{tabular}} & \multicolumn{2}{c}{trainval}             & \multicolumn{2}{c}{test}                 \\ \cline{2-5} 
                                                                           & AP(\%)                  & $C_{recall}$(\%)     & AP(\%)                  & $C_{recall}$(\%)     \\ \midrule
1                                                                          & 46.83                   & 5.60           & 40.35                   & 4.40           \\
2                                                                          & 90.75                   & 96.18          & 81.14                   & 94.80          \\
3                                                                          & \textit{\textbf{90.76}} & 96.43          & \textit{\textbf{89.19}} & 96.00          \\
4                                                                          & 90.60                   & 97.39          & 87.23                   & 96.40          \\
5                                                                          & 90.41                   & 97.44          & 80.62                   & 97.60          \\
$+\infty$                                                                          & 88.76                   & \textbf{97.73} & 75.57                   & \textbf{98.00} \\ \bottomrule
\end{tabular*}
\end{table}

As shown in Table \ref{tab:expand-num}, it achieves the best AP performance with expansion ratio 3 on the trainval set. When the expansion ratio is less than 3, the AP increases gradually; when the expansion ratio is greater than 3, the AP decreases gradually. Due to the restriction of the vehicle boundary, the expansion ratio $+\infty$ represents the vehicle region. As can be seen, a too small or too large expansion ratio leads to significant performance degradation, because a small region can not fully contain the license plate, and a large region is not favorable to small-sized license plate. Moreover, the $C_{recall}$ increases as the expansion ratio increases. However, when the expansion ratio is greater than 1, the $C_{recall}$ has little improvement, so we set it to 3 by default. The results on the test set further validate that an expanded region of appropriate size is favorable to license plate detection.

\begin{table*}[width=1.5\linewidth,cols=5,pos=h]
\caption{Comparative experiments (AP) with horizontal bounding box. "E2E" means end-to-end, "FC" means four corners, and "VP" means vehicle proposal.}\label{tab:comparative}
\begin{tabular*}{\tblwidth}{@{} LCCCC@{} }
\toprule
\multirow{2}{*}{Method} & \multicolumn{2}{c}{IOU=0.5}                                                                                      & \multicolumn{2}{c}{IOU=0.75}                                                                                     \\ \cline{2-5} 
                        & \begin{tabular}[c]{@{}c@{}}TILT720\\ Test\end{tabular} & \begin{tabular}[c]{@{}c@{}}TILT1080\\ Test\end{tabular} & \begin{tabular}[c]{@{}c@{}}TILT720\\ Test\end{tabular} & \begin{tabular}[c]{@{}c@{}}TILT1080\\ Test\end{tabular} \\ \midrule
Faster R-CNN \cite{DBLP:conf/nips/RenHGS15}            & 81.65\%                                                & 73.88\%                                                 & 13.63\%                                                & 14.29\%                                                 \\
YOLOv2 \cite{DBLP:conf/cvpr/RedmonF17}                  & 80.80\%                                                & 79.58\%                                                 & 51.66\%                                                & 49.32\%                                                 \\
SSD \cite{DBLP:conf/eccv/LiuAESRFB16}                     & 86.63\%                                                & 86.34\%                                                 & 47.06\%                                                & 53.88\%                                                 \\
\midrule
TextBoxes \cite{LiaoSBWL17}               & 69.67\%                                                & 67.56\%                                                       & 37.24\%                                                & 38.66\%                                                       \\
Method \cite{DBLP:conf/eccv/SilvaJ18}                & 74.67\%                                                      & 64.78\%                                                      & 42.67\%                                                       & 38.61\%                                                       \\
Method \cite{chen_simultaneous_2019}                    & 84.05\%                                                & 82.05\%                                                 & 45.35\%                                                & 53.42\%                                                 \\
\midrule
Ours (E2E+VP)       & 75.57\%                              & 74.29\%                               & 34.26\%                              & 35.52\%                               \\
Ours (E2E)       & \textit{\textbf{89.19\%}}                              & \textit{\textbf{87.67\%}}                               & \textit{\textbf{54.51\%}}                              & \textit{\textbf{56.92\%}}                               \\
\bottomrule
\end{tabular*}
\end{table*}

\begin{figure*}
    \centering
    \includegraphics[width=1.0\textwidth]{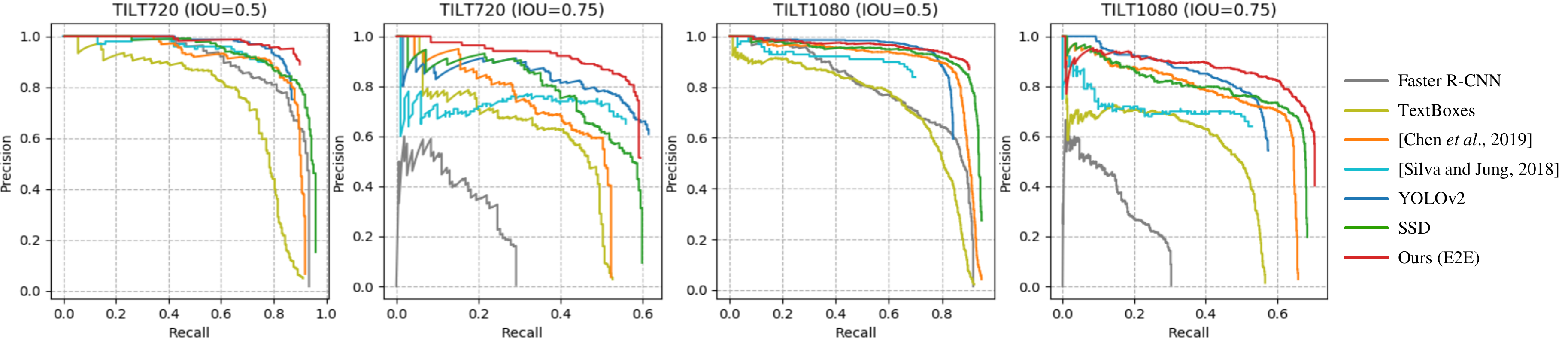}
    \caption{The precision-recall curve of different methods, datasets, and IOU thresholds. Our method achieves the best performance, especially for the larger IOU threshold.}
    \label{fig:pr-curve}
\end{figure*}

\subsection{Experiments with Horizontal Bounding Box}\label{sec4.5}

We compare Faster R-CNN \cite{DBLP:conf/nips/RenHGS15}, YOLOv2 \cite{DBLP:conf/cvpr/RedmonF17} and SSD \cite{DBLP:conf/eccv/LiuAESRFB16} with our proposed method, and the input size of all these methods is 512. The backbone network of Faster R-CNN, SSD, and our method is VGG-16 \cite{simonyan2015very}, while the backbone network of YOLOv2\footnote{The backbone network of YOLOv2 has 19 convolutional layers and is comparable to VGG-16.} is unchanged. Besides, we conduct comparative experiments with LPD methods \cite{DBLP:conf/eccv/SilvaJ18} and \cite{chen_simultaneous_2019} as well as scene text detection method TextBoxes \cite{LiaoSBWL17}. Except for \cite{DBLP:conf/eccv/SilvaJ18}\footnote{The authors have publicly released their trained models for license plate detection. Please refer to \url{https://github.com/sergiomsilva/alpr-unconstrained} for more details.}, all other methods are trained by ourselves with the trainval set of TILT720 and TILT1080 respectively. As shown in Table \ref{tab:comparative}, our method (E2E) achieves the best performance with different datasets and IOU thresholds. Moreover, as shown in Figure \ref{fig:pr-curve}, with a larger IOU threshold, our method (E2E) can obtain a larger performance gap than other methods. For example, with the IOU threshold 0.5, our method (E2E) is 2.56\% and 1.33\% better than SSD \cite{DBLP:conf/eccv/LiuAESRFB16} on the test set of TILT720 and TILT1080 respectively; with the IOU threshold 0.75, our method (E2E) is 7.45\% and 3.04\% better than SSD \cite{DBLP:conf/eccv/LiuAESRFB16}. Moreover, we enlarge the expanded region at the LREA stage to the whole vehicle region and utilize the vehicle proposal for the next MOLPR stage (E2E+VP). As shown in Table \ref{tab:comparative}, the detection performance is significantly degraded compared with detecting LP in the local region (E2E), because the vehicle proposal is too large and not favorable to small-sized license plate detection.

\begin{table*}[width=1.5\linewidth,cols=7,pos=h]
\caption{Comparative experiments with quadrilateral bounding box on the test set of TILT720. "E2E" means end-to-end, "FC" means four corners, and "VP" means vehicle proposal.}\label{tab:comparative-iblique}
\begin{tabular*}{\tblwidth}{@{} LCCCCCC@{} }
\toprule
\multirow{2}{*}{Method} & \multicolumn{3}{c}{IOU=0.5}                                     & \multicolumn{3}{c}{IOU=0.75}                                    \\ \cline{2-7} 
                        & P                & R                & F                        & P                & R                & F                        \\
\midrule
SSD \cite{DBLP:conf/eccv/LiuAESRFB16}                     & 98.66\%          & 58.80\% & 73.68\%                   & 65.10\%           & 38.80\%          & 48.62\%                   \\
Method \cite{DBLP:conf/eccv/SilvaJ18}                & 88.79\% & 76.00\%          & 81.90\%                   & 53.27\%          & 45.60\%          & 49.14\%                   \\
\midrule
Ours (E2E+VP)                    & 86.14\%          & 69.60\%          & 76.99\% & 45.05\% & 36.40\% & 40.27\% \\
Ours (SSD+FC)                    & 97.47\%          & 61.60\%          & 75.49\% & 75.32\% & 47.60\% & 58.33\% \\
Ours (E2E)                    & 90.61\%          & 88.80\%          & \textit{\textbf{89.70\%}} & 60.41\% & 59.20\% & \textit{\textbf{59.80\%}} \\
\bottomrule
\end{tabular*}
\end{table*}

\subsection{Experiments with Quadrilateral Bounding Box}\label{sec4.6}

\begin{figure*}
    \centering
    \includegraphics[width=1.0\textwidth]{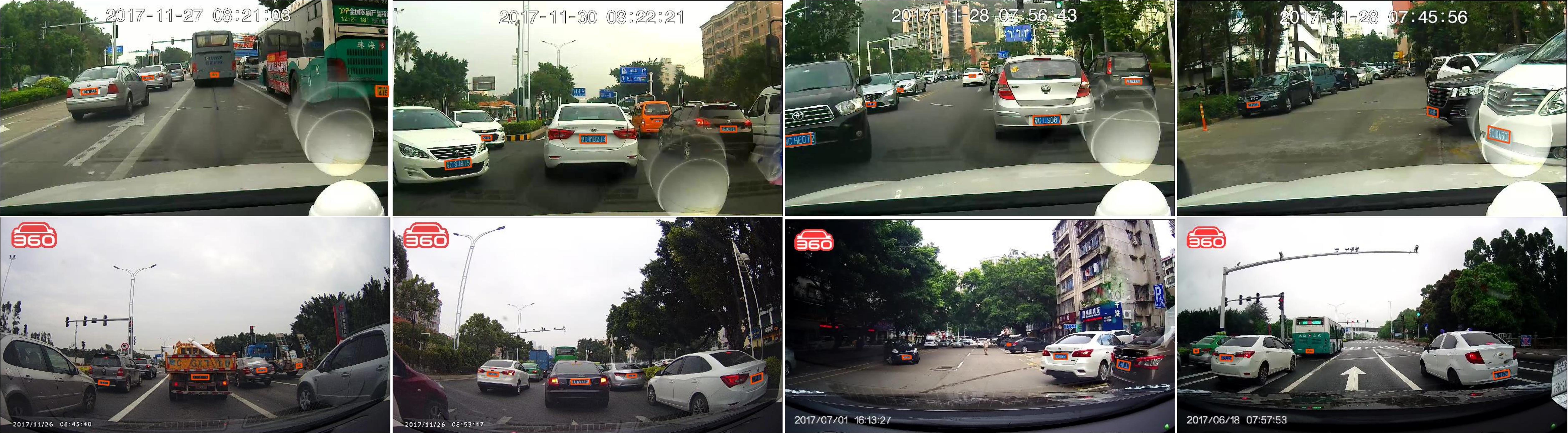}
    \caption{Detection results of TILT720 (first row) and TILT1080 (second row). Our method can correctly localize small-sized and oblique license plates as well as license plates of large buses and trucks.}
    \label{fig:results}
\end{figure*}

We conduct comparative experiments on the test set of TILT720 with different IOU thresholds. As shown in Table \ref{tab:comparative-iblique}, our method (E2E) achieves the best $F_1$-score with different IOU thresholds. SSD \cite{DBLP:conf/eccv/LiuAESRFB16} has poor performance because it can only detect the horizontal bounding box of the license plate. Furthermore, we upgrade vanilla SSD \cite{DBLP:conf/eccv/LiuAESRFB16} and make it capable of detecting the four corners of the license plate (SSD+FC), which simulates the MOLPR stage and can directly localize the horizontal and quadrilateral bounding box of the license plate in the input image. SSD+FC achieves better performance than vanilla SSD \cite{DBLP:conf/eccv/LiuAESRFB16}, especially for the larger IOU threshold. However, due to directly detecting LP from the large input image, SSD+FC still lags behind our end-to-end method (E2E). Besides, our method with vehicle proposal (E2E+VP) still suffers great performance degradation, which proves the effectiveness of detecting LP in the local region of our method (E2E). All methods, except for our method (E2E), suffer low recall, because these methods localize the license plate in a relatively large region (input image or vehicle region), which leaves the confidence of LP at a low level. Our method can localize the license plate in a small region around the license plate, which reduces background noises and can detect the license plate with high confidence. The aforementioned experiments prove that our method can precisely localize the quadrilateral bounding box of the oblique license plate. Some qualitative detection results are illustrated in Figure \ref{fig:results}.

\section{Conclusion}\label{sec5}
In this work, we propose an end-to-end trainable network for small-sized and oblique license plate detection via vehicle-plate relation mining, which detects the license plate in an end-to-end scheme. First, we propose a novel method to estimate the local region around the license plate using spatial relationships between the license plate and the vehicle, which can greatly reduce the search area and precisely detect very small-sized license plates. Second, we propose a new method to localize the quadrilateral bounding box by regressing the four corners of the license plate to robustly detect oblique license plates. Finally, based on the aforementioned methods, we develop an end-to-end trainable network for degraded license plate detection. Extensive experiments verify the effectiveness of our method, especially for a large IOU threshold. In future work, we would further promote the detection performance of vehicle detection to reduce false negatives of the license plate.

\printcredits

\bibliographystyle{cas-model2-names}

\bibliography{cas-refs}


\end{document}